\documentclass[11pt]{article}

\usepackage[final]{acl}

\usepackage{times}
\usepackage{latexsym}
\usepackage[T1]{fontenc}
\usepackage[utf8]{inputenc}
\usepackage{microtype}
\usepackage{inconsolata}
\usepackage[most]{tcolorbox}
\usepackage{wrapfig}
\usepackage{makecell}
\usepackage{multirow}
\usepackage{amsmath}
\usepackage{amssymb}
\usepackage{enumitem}
\usepackage[table]{xcolor}
\usepackage{graphicx}
\usepackage{tikz}
\usetikzlibrary{positioning,calc}
\tcbuselibrary{skins,breakable}

\definecolor{lightred}{RGB}{255, 180, 180}
\definecolor{lightgreen}{RGB}{180, 255, 180}
\definecolor{riskblue}{RGB}{186,216,240}
\definecolor{proborange}{RGB}{249,240,200}
\definecolor{lossgreen}{RGB}{219,231,231}
\definecolor{baseline}{RGB}{249,240,200}
\definecolor{frame1}{RGB}{111,135,141}
\definecolor{frame2}{RGB}{69,82,129}
\usepackage{booktabs}

\title{Rethinking Prospect Theory for LLMs: Revealing the Instability of Decision-Making under Epistemic Uncertainty}

\author{
Rui Wang\textsuperscript{1,4}\thanks{Equal contribution.},
~~Qihan Lin\textsuperscript{1,2}\footnotemark[1],
~~Jiayu Liu\textsuperscript{1,4}\footnotemark[1],
~~Qing Zong\textsuperscript{1},
~~Tianshi Zheng\textsuperscript{1}, \\
\bfseries\ Dadi Guo\textsuperscript{1},
~~Haochen Shi\textsuperscript{1},
~~Peixuan Han\textsuperscript{3},
~~Weiqi Wang\textsuperscript{1},
~~Yangqiu Song\textsuperscript{1}
\\
\textsuperscript{1}Hong Kong University of Science and Technology \\
\textsuperscript{2}Huazhong University of Science and Technology \\
\textsuperscript{3}University of Illinois Urbana-Champaign \\
\textsuperscript{4}HKUST-UGAIL \\
}

\begin{document}
\maketitle

\begin{abstract}
Real-world decision-making often involves uncertainty expressed in linguistic rather than numerical terms, and Prospect Theory (PT) provides a classic framework for modeling human behavior under such uncertainty. 
Although recent studies have developed frameworks to estimate PT parameters for Large Language Models (LLMs), few have examined whether PT itself adequately describes LLM decision-making behavior.
To address these gaps, we develop a streamlined workflow grounded in a classic behavioral economics experimental paradigm. 
First, we estimate PT parameters and evaluate how well the resulting model captures LLM decision-making behavior. 
We then derive probability mappings for epistemic markers in the same context and inject them into prompts to examine the stability of PT parameters under linguistic uncertainty.
Our findings suggest that PT does not consistently provide a reliable account of LLM decision-making across models, and that its application to LLMs is likely sensitive to epistemic uncertainty.
The findings caution against the deployment of PT-based frameworks in real-world applications where epistemic ambiguity is prevalent, giving valuable insights in behaviour interpretation and future alignment direction for LLM decision-making.
\footnote{Code: \url{https://github.com/HKUST-KnowComp/MarPT}.}
\end{abstract}

\section{Introduction}
\label{sec:intro}
LLMs are increasingly used in mission-critical decision-making tasks such as healthcare and finance~\citep{keith-stent-2019-modeling, lehman2022learningasklikephysician}.
While theoretical foundations for human decision-making under uncertainty are well-established, the inherent decision patterns and risk attitudes of LLMs are underexplored.
Among the psychological models used in the LLM field, Prospect Theory (PT)~\citep{Kahneman1979PT, Tanaka2010RiskMeasure} is particularly influential~\citep{horton2023large, Jia2024DMFrame}, and continues to play an important role in training, testing, and alignment frameworks for LLMs~\citep{cheng-etal-2025-weaponization, wang2025riskprofilingmodulationllms}. Despite its widespread adoption, the fundamental applicability of this human psychological model to LLMs is often assumed rather than systematically validated.

\begin{figure}[t]
     \centering
     \includegraphics[width=1\linewidth]{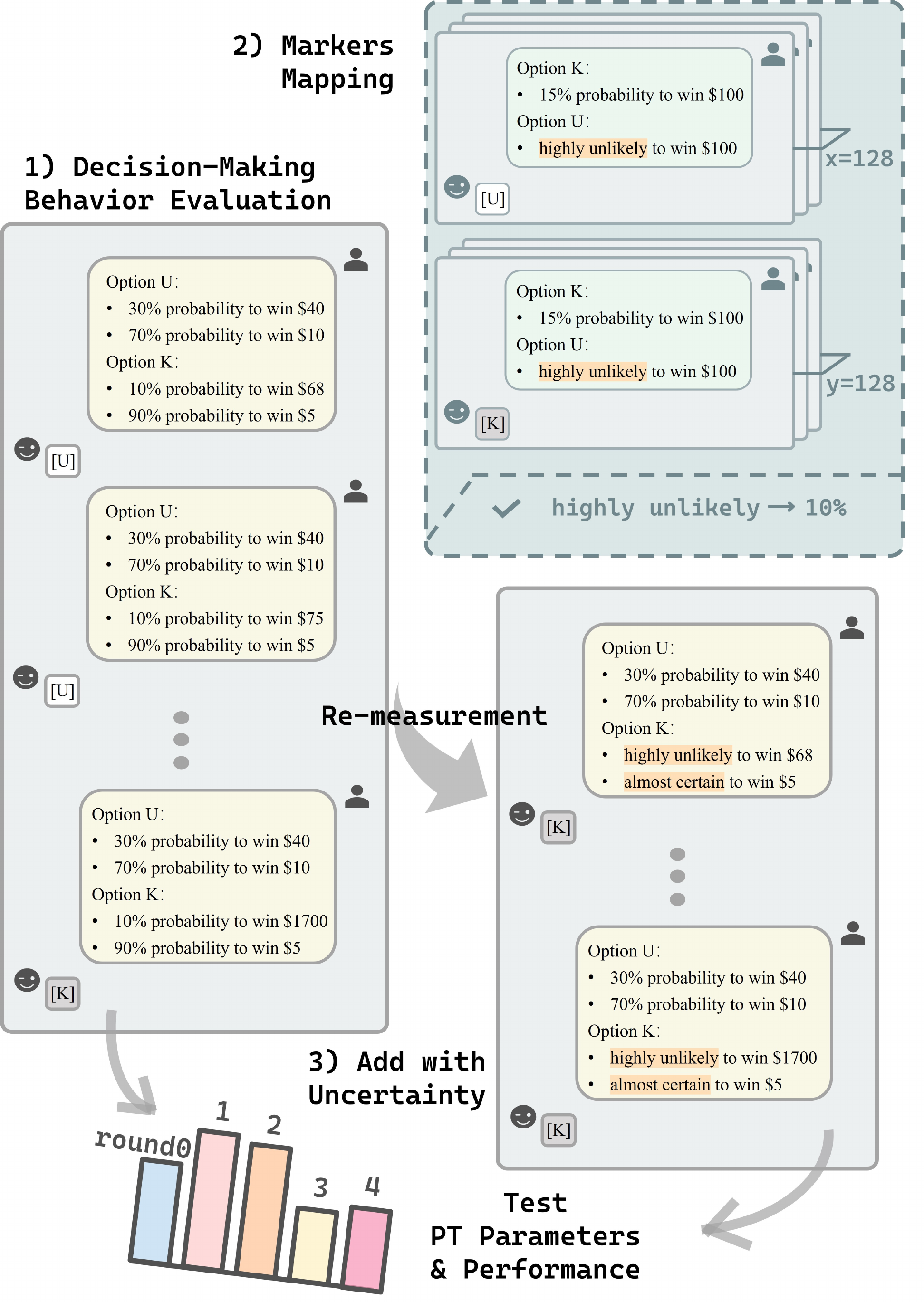}
     
     \caption{\textbf{An overview of our three-stage experiment.} 
    Stage~1 fits PT parameters from binary choices with precise probabilities. 
    Stage~2 estimates each marker’s internal probability from the point where options~K and~U are chosen equally. 
    Stage~3 substitutes probabilities with markers to measure the effect of linguistic uncertainty markers.
    }
    \label{fig:overview}
    \vspace{-0.23in}
\end{figure}

Beyond basic numerical evaluations, realistic decision-making involves pervasive linguistic uncertainty, predominantly expressed through epistemic markers~\citep{Belem2024perceptions, Liu2025markersinconf, Zhou2023navigating}.
Because humans naturally adopt such verbal expressions to communicate risk~\citep{wallsten1986measuring}, enabling models to process them is essential for robust human-AI communication~\citep{bhatt2021uncertainty}.
Given the highly safety-critical nature of AI-driven decision-making~\citep{leyliabadi2025conceptualframeworkaibaseddecision}, it is crucial that the theoretical frameworks we rely on to interpret and align LLM behavior remain reliable in real-world conditions. If Prospect Theory is to serve as a dependable lens for evaluating LLMs, its validity must not break down when uncertainty is expressed through epistemic markers.

However, whether Prospect Theory remains robust under linguistic uncertainty expressed by epistemic markers remains largely unexplored.
Existing studies have examined how personality prompting, sociodemographic embedding, role-playing interventions, or post-training can systematically shift LLMs' PT parameters~\citep{Jia2024DMFrame, Liu2025EandARisk, wang2025riskprofilingmodulationllms}, yet the question of how epistemic markers~\citep{lee2025llmjudgesrobustexpressionsuncertainty} influence LLM decision-making behavior under PT has been overlooked.

On this end, we design a lottery-based evaluation adapted from classic behavioral economics~\citep{Allais1953, Brandstatter2006priority, Tanaka2010RiskMeasure}, leveraging prior findings that epistemic markers can be mapped to numerical probabilities~\citep{Belem2024perceptions, zhou2023navigatinggreyareaexpressions, hu2025definedecisionmakinganalogicalreasoning}.
Our evaluation consists of three stages, illustrated in Figure~\ref{fig:overview}. 
We begin by presenting binary choice tasks with precise numerical probabilities to estimate each model’s PT parameters. 
In the second stage, LLMs make binary choices between numerical probabilities and epistemic markers.
This allows us to infer the model’s implicit probabilistic interpretation of those uncertainty expressions under similar contexts. 
Finally, we re-assess model behavior on the original decision tasks, now framed using epistemic markers grounded in their inferred probability values, to directly evaluate the impact of epistemic uncertainty on decision-making. 

Our results reveal two key insights. First, Prospect Theory does not consistently perform well at explaining LLM decision behaviors, and it exhibits model-wise differences where larger-scale models outperform the others.
Second, introducing epistemic markers disrupts decision consistency and alters PT parameters, exposing the fragility of LLMs' decision-making under linguistic uncertainty. 
The results suggest that LLMs do not exhibit universal Prospect-Theory-like behavior, and their risk tendencies are likely not robust under epistemic uncertainty.
It calls for specific explanation and alignment frameworks for more interpretable and robust LLM decision-making.
\section{Related Work}
\paragraph{Prospect Theory in Economics.}
Prospect Theory~\citep{Kahneman1979PT, Barberis2013ThirtyYears} has served as a foundational framework for modeling human decision-making under risk, capturing key behavioral patterns such as loss aversion and probability distortion.
Empirical studies have extended PT to diverse populations and settings~\citep{Tanaka2010RiskMeasure}, and recent work explores whether LLMs
exhibit similar patterns~\citep{Jia2024DMFrame, Liu2025EandARisk}.
Our work builds on these efforts but focuses on how linguistic uncertainty impacts PT-consistent behavior in LLMs.

\paragraph{LLM Decision-Making under Uncertainty.}
Recent work evaluates how LLMs handle economic decision-making and risk-sensitive choices, including whether their behavior follows prospect-theoretic patterns~\citep{Jia2024DMFrame}. 
Related studies further examine whether LLMs align with human economic risk preferences in risk assessment scenarios~\citep{Liu2025EandARisk}.
Beyond numerical risk settings, recent work also investigates whether epistemic markers can accurately reflect LLM uncertainty in confidence estimation~\citep{DBLP:conf/acl/Zong0ZRS25}.
More broadly, \citet{binz2025foundation} investigate whether a foundation model can predict and capture human behavior across a range of cognitive tasks.
We extend this line of inquiry by testing whether linguistic uncertainty conveyed by epistemic markers causes instability in LLM behaviors within decision-making contexts.

\paragraph{Epistemic Markers in LLMs.}
Recent work investigates how LLMs interpret and respond to linguistic signals related to uncertainty and confidence.
\citet{Zhou2023navigating} study how different epistemic markers embedded in prompts affect model predictions.
\citet{Belem2024perceptions} evaluate LLMs' interpretation of epistemic markers, finding partial human-like behavior alongside systematic biases.
\citet{Liu2025markersinconf} further argue that such markers are often unreliable indicators of internal confidence in LLMs.
These findings suggest that models may mimic surface linguistic patterns rather than demonstrate true epistemic reasoning.
Our study builds on this line of work by examining how LLMs process epistemic markers in uncertain economic contexts.
More comparative analysis with previous work are in Appendix \ref{app:comparative_analysis}.

\section{Preliminaries}

\begin{figure*}[htbp]
\centering
\resizebox{\linewidth}{!}{
\begin{tikzpicture}[font=\normalsize, node distance=0.8cm]

\node[minimum width=1.5cm, minimum height=2.6cm, 
      fill=riskblue!80!white, name=risk] {};
\fill[riskblue!40!white] (risk.south west) rectangle (risk.south east |- 0,-0.5);
\node[black] at ([yshift=-0.3cm]risk.north) {+$\infty$};
\node[black] at (risk.center) {1};
\node[black] at ([yshift=0.3cm]risk.south) {0};

\node[right=0.7cm of risk.north, yshift=-0.18cm, anchor=west] 
  {$\uparrow$ Risk-seeking};
\node[right=0.7cm of risk.center, anchor=west] 
  {$\rightarrow$ Risk neutral};
\node[right=0.7cm of risk.south, yshift=0.18cm, anchor=west] 
  {$\downarrow$ Risk-averse};

\node[right=3cm of risk.east, 
      minimum width=1.5cm, minimum height=2.6cm, 
      fill=lossgreen, name=loss] {};
\fill[lossgreen!40!white] (loss.south west) rectangle (loss.south east |- 0,-0.5);
\node at ([yshift=-0.3cm]loss.north) {+$\infty$};
\node at (loss.center) {1};
\node at ([yshift=0.3cm]loss.south) {0};

\node[right=0.7cm of loss.north, yshift=-0.18cm, anchor=west] 
  {$\uparrow$ More sensitive to loss};
\node[right=0.7cm of loss.center, anchor=west] 
  {$\rightarrow$ Neural evaluation};
\node[right=0.7cm of loss.south, yshift=0.18cm, anchor=west] 
  {$\downarrow$ More sensitive to gain};

\node[right=3.9cm of loss.east, 
      minimum width=1.5cm, minimum height=2.6cm, 
      fill=proborange, name=prob] {};
\fill[proborange!45!white] (prob.south west) rectangle (prob.south east |- 0,-0.5);
\node at ([yshift=-0.3cm]prob.north) {+$\infty$};
\node at (prob.center) {1};
\node at ([yshift=0.3cm]prob.south) {0};

\node[right=0.7cm of prob.north, yshift=-0.18cm, anchor=west] 
  {$\uparrow$ Underweighting small probabilities};
\node[right=0.7cm of prob.center, anchor=west] 
  {$\rightarrow$ No probability distortion};
\node[right=0.7cm of prob.south, yshift=0.18cm, anchor=west] 
  {$\downarrow$ Overweighting large probabilities};

\node[above=0.25cm, xshift=0.18cm] at (risk.north) {\textbf{$\sigma$: Risk Preference}};
\node[above=0.25cm, xshift=0.18cm] at (loss.north) {\textbf{$\lambda$: Loss Aversion}};
\node[above=0.20cm, xshift=0.8cm] at (prob.north) {\textbf{$\gamma$: Probability Weighting}};

\end{tikzpicture}
}
\caption{\textbf{Visual illustration of the three PT parameters ($\sigma$, $\lambda$, $\gamma$).} Each parameter is shown with its meaning and an interpretation of its directional significance.}
\vspace{-0.1in}
\label{fig:three_parameters}
\end{figure*}

In rational decision theory, an agent's preference over risky prospects follows the \textit{von Neumann-Morgenstern expected utility framework}~\citep{neumann1944theory}.
For a prospect $P = (x_1, p_1; \cdots; x_n, p_n)$ yielding outcome $x_i$ with probability $p_i$, the \textit{expected utility} is computed as:
\begin{equation}
EU(P) = \sum_{i=1}^{n} p_i \cdot u(x_i),
\end{equation}
where $u(\cdot)$ is a cardinal utility function mapping outcomes to real numbers. 
Under traditional Expected Utility Theory (EUT)~\citep{vonNeumann1944}, human decisions are assumed to maximize $EU(P)$.
However, empirical evidence systematically violates EUT assumptions. 
Prospect Theory (PT) addresses these anomalies through three psychological distortions of rational utility.
It maintains a utility-based approach but fundamentally alters Expected Utility Theory through three key properties:
\begin{itemize}[leftmargin=*, itemsep=2pt, parsep=0pt]
    \item \textbf{(1) Risk Preference ($\sigma$)}: Agents often exhibit risk aversion or risk-seeking behavior.
    \item \textbf{(2) Loss Aversion ($\lambda$)}: Losses psychologically outweigh equivalent gains.
    \item \textbf{(3) Probability weighting ($\gamma$)}: Agents often exhibit systematic probability distortion.
\end{itemize}
To capture these characteristics, Prospect Theory introduces two specialized functions: the \textit{value function} formalizes how outcomes translate into subjective utility, while the \textit{probability weighting function} captures non-linear probability perception. 
Together, these functions model the distorted utility calculations that characterize PT decision-making.

The \textbf{value function} $v(x)$ quantifies subjective satisfaction from outcomes relative to a reference point (zero in this study). For outcomes $x\geq0$ (gains) and $x<0$ (losses), the value function is:

\begin{equation}
v(x) = 
\begin{cases}
x^\sigma & \text{for } x\geq0 \\
-\lambda(-x)^\sigma & \text{for } x<0.
\end{cases}
\end{equation}

Key parameters here are loss aversion ($\lambda$) which is a multiplier of  negative perceptions, and risk preference ($\sigma$) which controls curvature (sensitivity to values).

The \textbf{probability weighting function} formalizes the transformation of objective probabilities $p$ into subjective decision weights:

\begin{equation}
    w(p) = \frac{p^\gamma}{\left(p^\gamma + (1-p)^\gamma \right)^{1/\gamma}},
\end{equation}
where $\gamma$ controls the curvature of the function.

The final utility for any binary prospect in the form $P = (x, p; y, q)$ is defined as follows:
\begin{equation}
u(P) =
\begin{cases}
v(y) + w(p)(v(x) - v(y)), & \text{[1]} \\
w(p)v(x) + w(q)v(y),      & \text{[2]}
\end{cases}
\end{equation}

where [1] denotes $x > y > 0$ or $x < y < 0$, and [2] denotes $x < 0 < y$. All functions follow the standard formulations set in~\citep{Kahneman1979PT}.

\section{Decision-Making Behavior Evaluation}

\paragraph{Experimental setup.}
Based on the above economic frameworks, we adopt the three-series lottery-choice experiment developed by~\citet{Tanaka2010RiskMeasure} to get a reliable PT parameter measurement.
Series 1 and 2 are designed to elicit risk preference ($\sigma$) and probability weighting ($\gamma$), while series 3 is designed for loss aversion ($\lambda$). 
The prospect settings are shown in Appendix \ref{app:prospect-settings}, and the prompt design is in Appendix \ref{app:prompt-design}.
The experiment setup hyperparameters are summarized in Appendix~\ref{app:experiment-hyperparameters}.

Each lottery consists of two options: a safe option K and a riskier option U. The agent is asked to directly choose between options K and U based on its risk preference. 
After sampling $\eta$ times for each question, we count the portion of choosing option K for each lottery.
Then we define the predicted probability of choosing option K for each lottery as follows:
\begin{equation}
P(\text{choose K}) = \frac{e^{\Delta EU}}{1 + e ^ {\Delta EU}},
\end{equation}
where ${\Delta EU} = u(\text{K}) - u(\text{U})$ is the difference in the distorted utility of prospect K and U under prospect theory.
We choose the sigmoid function as it is a standard formulation in economic studies~\citep{chakravarty2009recursive}.

\paragraph{PT Parameters and Confidence Intervals.}
We add up the Bernoulli log-likelihood for all $M$ lotteries as the negative log-likelihood function, and run Maximum Likelihood Estimation (MLE) with this function to estimate $\sigma, \lambda \text{ and } \gamma$.
To get the confidence intervals, we use a bootstrap method~\citep{Efron1979Bootstrap} by generating simulated datasets through binomial sampling from predicted probabilities derived from the original parameter estimates. Specifically, for each observation $i$, we sample $\tilde{y}_i \sim \text{Binomial}(n=1, p=\hat{p}_i)$ where $\hat{p}_i$ is the predicted probability. The model parameter standard deviation $\sigma_{\hat{\theta}}$ is estimated from the bootstrap distribution, and the $(1-\alpha)$ confidence interval is constructed using the percentile method:
\begin{equation}
    \text{CI}_{1-\alpha} = \left[ \hat{\theta}_{(\alpha/2)}^*, \hat{\theta}_{(1-\alpha/2)}^* \right],
\end{equation}
where $\hat{\theta}_{(\alpha)}^*$ denotes the $\alpha$-quantile of the bootstrap parameter estimates. This approach accounts for parameter uncertainty in finite samples.

\paragraph{PT Goodness-of-fit.} We then quantify model goodness-of-fit by computing \textbf{McFadden pseudo-$\mathbf{R^2}$}~\citep{mcfadden1977}, defined as:
\begin{equation}
    R^2_{\text{McFadden}} = 1 - \frac{\mathcal{L}_{\text{PT}}}{\mathcal{L}_{\text{null}}},
\end{equation}
where $\mathcal{L}_{\text{PT}}$ is the log-likelihood of Prospect Theory and $\mathcal{L}_{\text{null}}$ represents the log-likelihood of the intercept-only model with uniform choice probabilities. This metric measures the improvement of our model over random guessing.
Finally, we calculate the \textbf{mean absolute error} (MAE) between the actual probability $p_{\text{actual}}$ and the predicted probability $p_{\text{pred}}$ of choosing option K derived from our PT model:
\begin{equation}
    \text{MAE} = \frac{1}{N} \sum_{i=1}^{N} \left| p_{\text{actual}}^{(i)} - p_{\text{pred}}^{(i)} \right|,
\end{equation}
where $N$ denotes the number of observations. This provides a direct measure of prediction error.

\paragraph{Summary.} Overall, the three PT parameters, their confidence intervals, the McFadden pseudo-$R^2$, and the MAE score together describe the revealed agent's risk attitude and our descriptive model's explanatory power.
\section{Evaluation with Epistemic Uncertainty}
Real-world decisions are often made under vague linguistic uncertainty rather than precise numerical probabilities~\citep{wallsten1986measuring, Belem2024perceptions}. This necessitates experiments to understand how epistemic markers influence LLMs' decision-making behavior.
We investigate how decision-making is affected when numerical probabilities are replaced by verbal probability expressions, or epistemic markers.
Section~\ref{sec:Probability-Mapping-of-Epistemic-Markers} estimates their numerical equivalents via a controlled lottery experiment, and Section~\ref{sec:Re-measurement-of-Prospect-Theory-Parameters} applies these values in the PT framework to re-measure PT parameters. 
The prompt design is in Appendix \ref{app:prompt-design}.

\begin{table}[h]
  \centering
  \small
  \begin{tabular}{lcc}
    \toprule
    \textbf{No.} & \textbf{Epistemic Marker} & \makecell[c]{\textbf{Probability Mapping} \\ \textbf{by Human}} \\
    \midrule
    1   & \textit{almost certain} & 95\% \\
    2   & \textit{highly likely} & 90\% \\
    3   & \textit{very likely} & 90\% \\
    4   & \textit{likely} & 80\% \\
    5   & \textit{probable} & 70\% \\
    6   & \textit{somewhat likely} & 70\% \\
    7   & \textit{possible} & 60\% \\
    8   & \textit{uncertain} & 50\% \\
    9   & \textit{somewhat unlikely} & 30\% \\
    10  & \textit{unlikely} & 25\% \\
    11  & \textit{not likely} & 20\% \\
    12  & \textit{doubtful} & 20\% \\
    13  & \textit{very unlikely} & 10\% \\
    14  & \textit{highly unlikely} & 10\% \\
    \bottomrule
  \end{tabular}
  \vspace{-0.05in}
  \caption{Epistemic markers and human probability mappings from~\citet{Belem2024perceptions}.}
  \label{marker}
  \vspace{-0.2in}
\end{table}

\subsection{Marker Probability Mappings}
\label{sec:Probability-Mapping-of-Epistemic-Markers}

\paragraph{Motivation.} Epistemic markers are inherently vague and context-sensitive~\citep{Liu2025markersinconf, Bergqvist+2015+123+141}, yet they often substitute for precise numerical probabilities in practice~\citep{Belem2024perceptions, zhou2023navigatinggreyareaexpressions, hu2025definedecisionmakinganalogicalreasoning}.
For LLMs, the ability to interpret epistemic markers consistently and meaningfully is critical if they are used as decision-support tools. 

\paragraph{Experimental setup.} To empirically examine how LLMs interpret these markers, and whether their interpretations are coherent and aligned with human intuition, we design a controlled lottery experiment in an economic decision-making context.
Each trial presents the model with a choice between two options.
For option K, there is a fixed probability $p$\% of winning \$$M$, while $p$ ranges over all values in set $probs$ (see Appendix \ref{app:hyperparameters}).
For option U, there is an unknown probability of winning \$$M$ which is defined by the different markers. 
For the usage of epistemic markers, We manually select 14 markers commonly used in prior work~\citep{Belem2024perceptions} to ensure that they are suitable for this context (see Table~\ref{marker}).

\begin{figure}[h]
     \centering
     \includegraphics[width=1\linewidth]{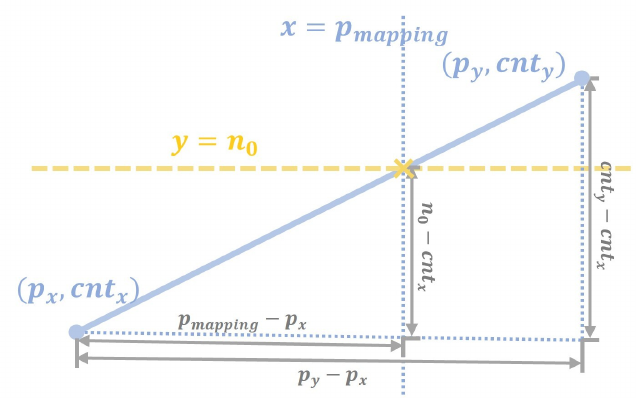}
     \caption{\textbf{Illustration of calculating $p_{mapping}$.} We use linear interpolation via similar triangles.}
     \label{fig:mapping_fig}
\end{figure}

\paragraph{Probability interpolation.} For each marker, we record the number of times the model selects option K, denoted as $\text{NUM}_K$.
The key assumption is that when $\text{NUM}_K$ reaches half of the total trials (i.e., an implied probability $p_0=0.5$), the model considers the two options equally attractive. We define the inferred probability mapping, $p_{\text{mapping}}$, for that marker as the value of $p$ where this equilibrium occurs.
Since $p$ is sampled at discrete points, the exact $p_0$ point may fall between two sampled probabilities. 
To estimate $p_{\text{mapping}}$, we perform linear interpolation between the two nearest points.
Let $n_0$ be the target count (i.e., $50\%$ of the sample size). Let $(p_x, \text{cnt}_x)$ and $(p_y, \text{cnt}_y)$ be the probability-count pairs immediately below and above $n_0$, respectively.
As illustrated in Figure \ref{fig:mapping_fig}, by the slope formula
\begin{equation}
    \frac{p_{mapping} - p_x}{p_y - p_x} = \frac{n_0 - cnt_x}{cnt_y - cnt_x},
\end{equation}
we solve for $p_{mapping}$:
\begin{equation}
\label{mapping2}
    p_{mapping} = \frac{(n_0 - cnt_x)\cdot p_y + (cnt_y - n_0)\cdot p_x}{cnt_y - cnt_x}.
\end{equation}
Through this experiment, we obtain a list of 14 probability values (one per marker) for each model, capturing how it semantically interprets verbal uncertainty in economic terms.

\begin{table}[h]
\centering
\small
\resizebox{\linewidth}{!}{
\begin{tabular}{lcccccccccccc}
\toprule
\textbf{Model} & \textbf{$\sigma$} & \textbf{$\lambda$} & \textbf{$\gamma$} & \textbf{MAE}$\downarrow$ & \textbf{R$^2$}$\uparrow$ \\
\midrule
\textit{Human}        & 0.670  & 2.630  & 0.685  & -  & - \\
\textit{Llama-3.1-8B-Instruct}   & 0.585  & 0.010  & 0.753  & 0.332  & 0.092 \\
\textit{Mistral-7B-Instruct-v0.3}& 0.534  & 0.570  & 0.577  & \textbf{0.155}  & \textbf{0.132} \\
\textit{Qwen2.5-7B-Instruct}     & 0.429  & 0.010  & 3.645  & \textbf{0.047}  & \textbf{0.116} \\
\textit{Qwen2.5-14B-Instruct}    & 0.503  & 1.909  & 0.896  & 0.257  & 0.067 \\
\textit{Qwen2.5-32B-Instruct}    & 0.598  & 1.213  & 0.867  & \textbf{0.161}  & \textbf{0.225} \\
\midrule
\textit{GPT-5-Mini}    & 0.447  & 4.000  & 2.888  & \textbf{0.115}  & \textbf{0.206} \\
\textit{Gemini-2.5-Flash}    & 0.495  & 1.499  & 0.814  & \textbf{0.134}  & \textbf{0.225} \\
\bottomrule
\end{tabular}
}
\caption{\textbf{Baseline Prospect Theory (PT) parameter estimation across evaluated LLMs.} Human benchmarks are sourced from~\citep{Tanaka2010RiskMeasure}. For reasoning-capable models (\textit{GPT-5-Mini} and \textit{Gemini-2.5-Flash}), we employ chain-of-thought prompts while explicitly prohibiting expected-value calculations. Bolded entries denote a reliable fit based on the criteria of $MAE \leq 0.20$ and $R^2 \geq 0.10$.
}
\label{tab:pt-parameters}
\end{table}

\subsection{Re-measurement of PT Parameters}
\label{sec:Re-measurement-of-Prospect-Theory-Parameters}
\paragraph{Marker Normalization.} After establishing the probability mappings, we select a pair of epistemic markers whose normalized probabilities approximate the original numerical settings, ensuring the core lottery structure remains consistent~\citep{budescu1995processing}.
Detailed replacement rules are provided in Appendix \ref{marker-replacement}. 
For example, for a model with \textit{``somewhat unlikely''} mapped to $ p_1 = 32\% $ and \textit{``highly likely''} mapped to $ p_2 = 68\% $, 
we normalize them as:
\begin{equation}
p'_1 = \frac{p_1}{p_1 + p_2}, \quad 
p'_2 = \frac{p_2}{p_1 + p_2}
\end{equation}
This normalization follows the standard approach of converting membership values to valid probability distributions~\citep{budescu1995processing, wallsten1986measuring}.

\paragraph{Lottery with Markers.} We replace the probabilities in the original decision-making behavior evaluation framework with the closest epistemic marker pairs, and re-run the PT measurement test using $p'_1$ and $p'_2$ and compare the result with the original study.
We design four experimental rounds to incrementally introduce linguistic uncertainty: Round 1 introduces markers only to Option K in Series 1 and 2; Round 2 to Option K across all three series; Round 3 to Option U only; and Round 4 to both options across all series.
This design enables systematic investigation of how epistemic markers affect decision-making across safe versus risky options (Option K versus U) and gain versus loss domains (Series 1-2 versus Series 3).
\section{Results and Findings}
\subsection{Numerical Probabilities Setting}

For decision-making under exact probability, we first estimate Prospect Theory parameters using the numerical lottery settings as the baseline condition.
The parameter values are shown in Table~\ref{tab:pt-parameters}, with additional details provided in Appendix~\ref{app:detail-results}.

\paragraph{Prospect Theory does not uniformly explain LLM decision behavior.}
Before interpreting the estimated PT parameters, we first examine whether each model's choice behavior can be meaningfully captured by the Prospect Theory framework. 
Following human econometric model regression standards, we consider an MAE score $> 0.20$ to indicate an unreliable regression result, and a McFadden pseudo-$R^2$ score $< 0.10$ to imply that applying PT lacks explanatory power~\citep{mcfadden1977}. 
Under these criteria, \textit{Llama-3.1-8B-Instruct} and \textit{Qwen2.5-14B-Instruct} exhibit both high prediction error and weak likelihood improvement over the null model, indicating that their choices cannot be reliably summarized by a stable PT parameterization. 
This suggests that the failure is not merely a matter of obtaining different $\sigma$, $\lambda$, or $\gamma$ values, but reflects a deeper mismatch between their observed choice patterns and the structural assumptions of Prospect Theory. 
This highlights the importance of evaluating model fit before attributing human-like economic preferences to LLMs, since poor PT fit may reveal unstable or internally inconsistent decision behavior rather than interpretable risk attitudes.

\paragraph{Larger LLMs exhibit more Prospect-Theory-like behavior, but loss aversion remains unstable.}
Overall, larger and more capable LLMs show stronger alignment with Prospect Theory, as reflected by higher McFadden pseudo-$R^2$ and lower MAE, while their estimated parameters still reveal an incomplete form of human-like decision behavior.
Across models with reliable PT fits, the risk preference parameter $\sigma$ mostly falls between 0.4 and 0.6, suggesting a consistent degree of risk aversion that is slightly weaker than that typically observed in humans.
However, the loss aversion parameter $\lambda$ is much less stable: some models are more sensitive to losses ($\lambda > 1$), whereas others appear more sensitive to gains ($\lambda < 1$).
In particular, \textit{Llama-3.1-8B-Instruct} and \textit{Qwen2.5-7B-Instruct} yield unusual boundary values of $\lambda$, indicating irregular choice patterns rather than stable reference-dependent preferences.
Together with the improved fit of larger models, these results suggest that PT-aligned behavior may partially emerge with model scale, but current LLMs mainly reproduce surface-level risk heuristics without consistently encoding the reference-dependent value structure underlying human loss aversion.

\begin{figure*}[htbp]
     \centering
     \hspace*{-1cm}
     \includegraphics[width=1\linewidth,height=6.7cm]{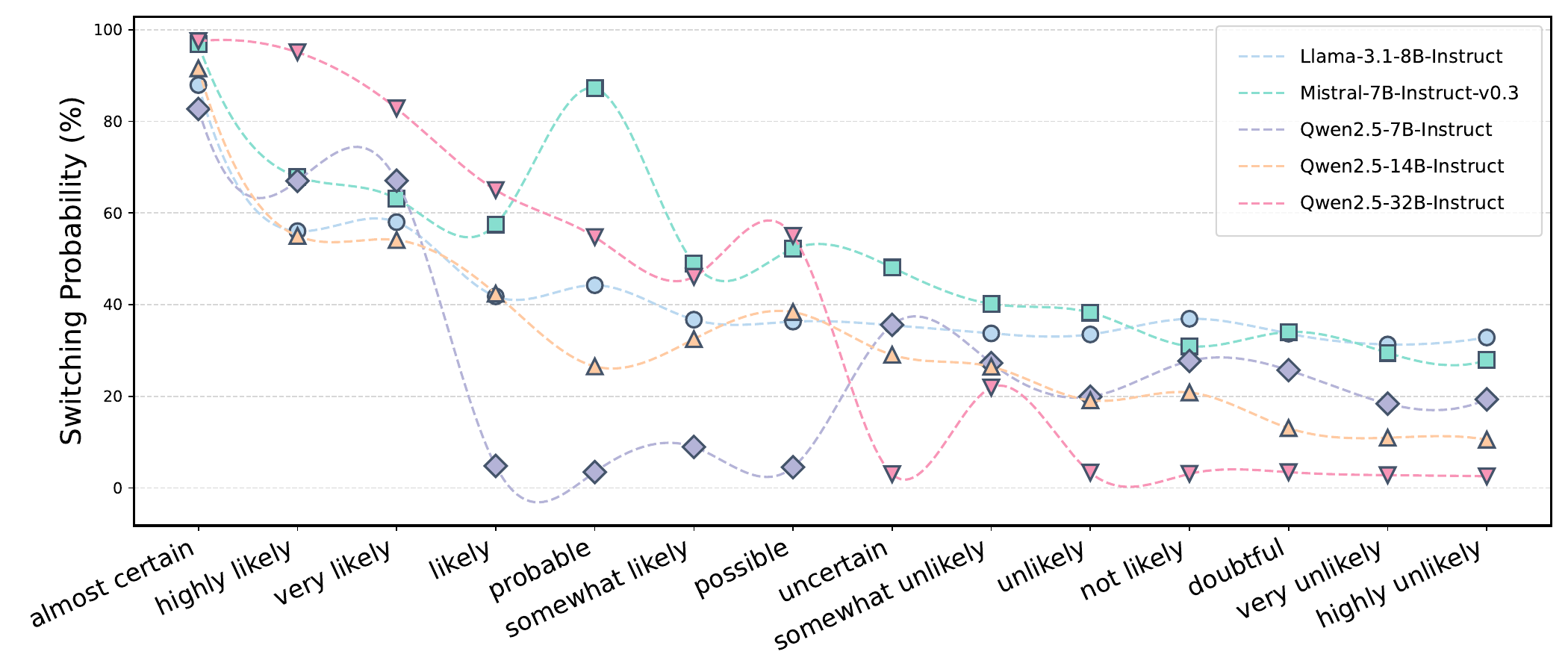}
     \caption{Probability mapping of different models for different markers.}
     \label{fig: show_markers}
\end{figure*}

\subsection{Epistemic Marker Mappings}

The detailed experimental results are shown in Tables Appendix~\ref{app:detail-results}. We also further visualize the probability mapping of different models for different epistemic markers in Figure~\ref{fig: show_markers}.

\paragraph{Epistemic marker mappings show reliable relative ordering across models.}
The probability mappings exhibit a consistent relative ranking across models, indicating that our mapping procedure captures stable ordinal semantics rather than noisy numerical artifacts.
As shown in Figure~\ref{fig: show_markers}, high-certainty markers such as \textit{`almost certain''}, \textit{`highly likely''}, and \textit{``very likely''} are consistently assigned among the highest probabilities, while low-certainty markers are mapped to substantially lower values.
Although the absolute probabilities vary across models, this stable ordering suggests that LLMs largely agree on the relative strength of epistemic markers.
This consistency also further reduces concerns that the mappings reflect model-specific noise.
This provides a reliable foundation for our subsequent analysis.

\paragraph{Model-specific mappings are approximately aligned with human probability judgments.}
The learned mappings are broadly consistent with the human mappings reported in prior work, supporting their external validity.
As shown in Table~\ref{marker}, most epistemic markers fall within the 5
At the same time, the mappings are not identical across models, as Figure~\ref{fig: show_markers} shows model-dependent differences in the absolute probability assigned to the same marker.
These differences indicate that a single fixed mapping would obscure meaningful variation in how models interpret the same epistemic expressions.
Therefore, our mapping procedure provides probability estimates that are both reliable and model-specific: they preserve the broad human-aligned semantics of epistemic language while capturing how each model uniquely interprets linguistic uncertainty.

\subsection{PT under Epistemic Markers}

\begin{figure*}[t]
     \centering
     \includegraphics[width=1\linewidth, height=6cm]{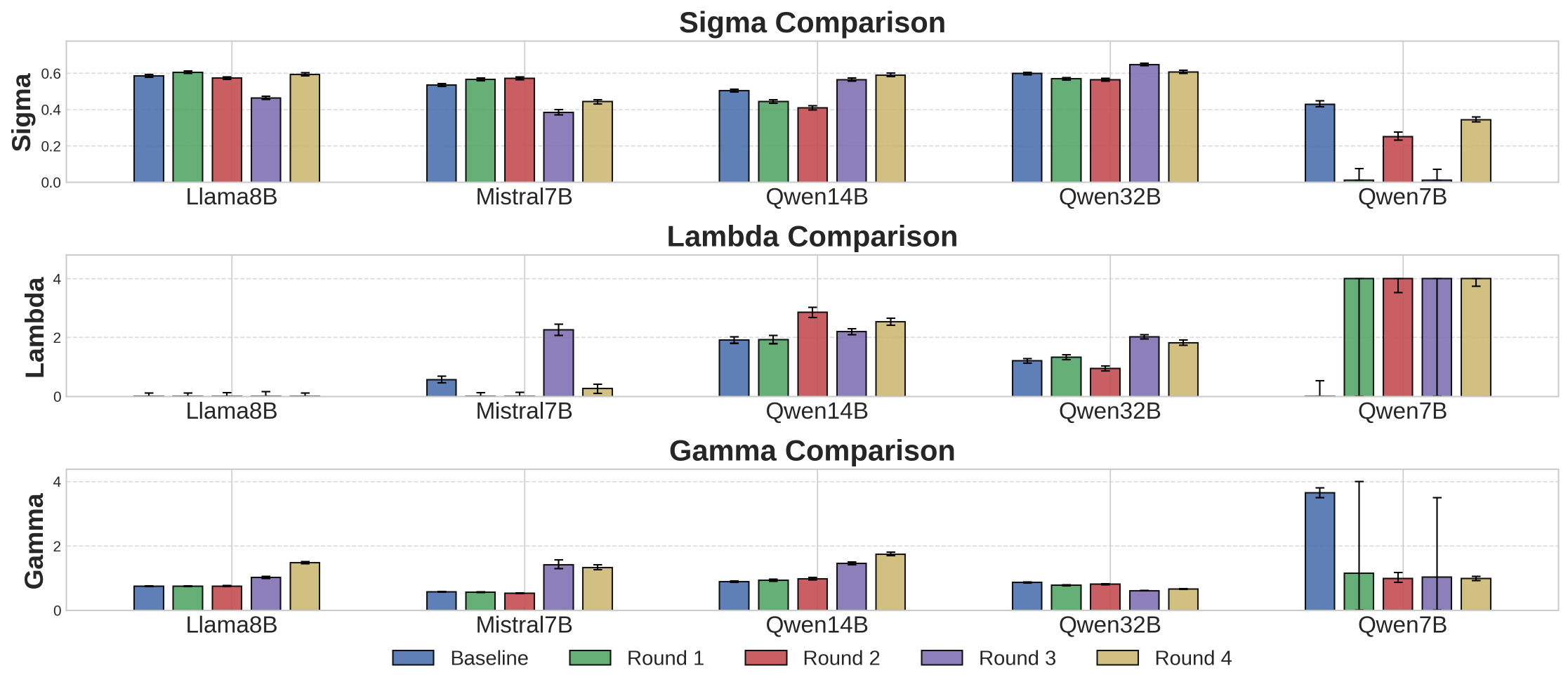}
     \caption{\textbf{PT parameters estimated across rounds.} Baseline corresponds to the first stage; subsequent rounds show parameter fluctuations.}
    \label{fig:parameters}
\end{figure*}

\begin{figure*}[htbp]
     \centering
     \includegraphics[width=1\linewidth]{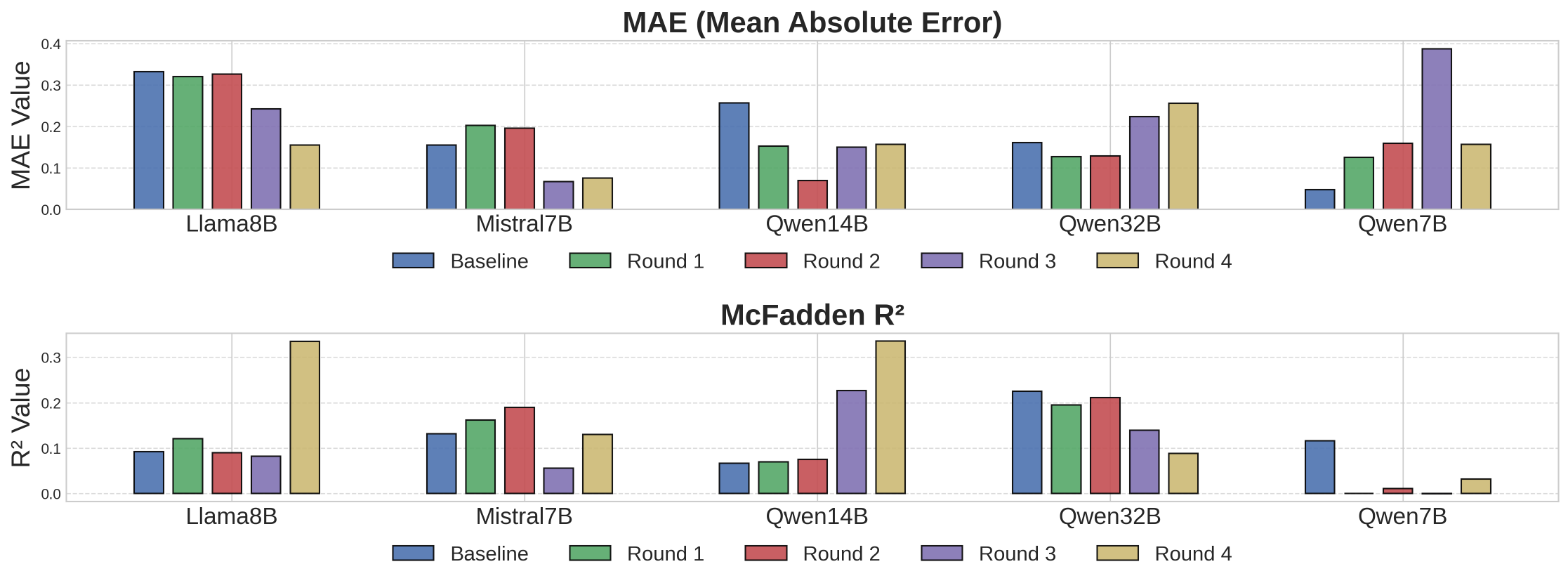}
     \caption{\textbf{PT performance across rounds.} Baseline represents initial PT fitness; MAE and McFadden $R^2$ vary considerably in later rounds.}
    \label{fig:performance}
\end{figure*}

The estimation results under varying degrees of linguistic uncertainty are illustrated in Figures~\ref{fig:parameters} and~\ref{fig:performance}. 
We examine how replacing precise numerical probabilities with epistemic markers affects LLMs' internal decision-making parameters and performance under the framework of Prospect Theory. 
Because the epistemic markers are selected based on the model-specific probability mappings established above, this stage further tests whether semantically matched linguistic uncertainty preserves the same revealed decision-making behavior as numerical probability.
The analysis covers four experimental conditions, including marker substitution in either or both options of the lottery choice task.

\paragraph{Epistemic uncertainty induces non-invariant PT parameters under matched probability meanings.}
For the majority of models, the estimated risk preference remains relatively stable despite replacing numeric probabilities with epistemic markers, indicating that risk attitudes are only mildly perturbed by uncertainty.
However, loss aversion and probability weighting shift more substantially under these conditions.
In particular, changes in $\lambda$ suggest that models do not preserve the same loss sensitivity when equivalent uncertainty is expressed through linguistic markers rather than numerical probabilities.
Similarly, the increase in $\gamma$ after adding disturbance shows that most models exhibit increasingly conservative probability distortions, reflecting cautious decision patterns under ambiguity.
More fundamentally, drift under semantically matched substitutions reveals non-invariant decision-making: models alter revealed preferences when equivalent uncertainty is expressed differently.
These shifts indicate that semantic equivalence in probability does not guarantee behavioral equivalence in model decisions.
This suggests that epistemic markers are not merely interchangeable verbal forms of numerical probabilities, but can directly affect the internal preference structure revealed by LLMs.

\paragraph{Larger size LLMs have more stable but still varied decision-making behavior.}
Model scale is associated with greater robustness to epistemic uncertainty, but not with full stability. 
Smaller language models, such as \textit{Mistral-7B-Instruct-v0.3}, exhibit substantial fluctuations in PT parameters across rounds after numerical probabilities are replaced with epistemic markers. 
In some conditions, their estimates even reach boundary values, indicating an unstable fit. 
By contrast, larger models such as \textit{Qwen2.5-32B-Instruct} show comparatively more stable parameter estimates under the same perturbations. 
However, their parameters still vary noticeably across experimental conditions, suggesting that increased model scale improves robustness to linguistic uncertainty without eliminating the sensitivity of decision-making behavior to epistemic phrasing.
This indicates that scale may mitigate, but does not resolve, the behavioral gap between numerical and linguistic representations of uncertainty. This scaling pattern suggests that robustness to epistemic phrasing improves gradually with model capacity.

\paragraph{Epistemic markers have model-dependent effects on PT fitness.}
While epistemic markers generally worsen PT performance metrics, their effects are not uniformly negative across models and rounds.
For example, in the four rounds with epistemic markers, MAE rises significantly for \textit{Qwen2.5-7B}, but it drops for \textit{Qwen2.5-14B}.
This mixed pattern suggests that epistemic uncertainty interacts with model-specific decision mechanisms and that conclusions about PT fitness may depend on how uncertainty is expressed.
This variation cautions against drawing overly broad model-agnostic conclusions about the effect of epistemic markers on PT fitness.
Given that epistemic markers are common in real-world communication, this model-dependent instability highlights the need to evaluate LLM decision-making under linguistic uncertainty rather than relying only on exact numerical probability settings.

\section{Discussions}
\label{sec:discussion}
\paragraph{On the Contributions and Significance of Our Work.}
Prospect Theory (PT) is a widely used framework for describing human decisions under risk \citep{Kahneman1979PT,Barberis2013ThirtyYears}.
Recent studies have applied PT to LLMs, but they have reached different conclusions about how well it describes model behavior \citep{Jia2024DMFrame,Liu2025EandARisk,payne2025analysis}.
Our results clearly show that the reliability of PT depends on both the model being studied and the way uncertainty is expressed.
Under numerical probabilities, PT meets our fit criteria for some models but not for others. Even when the numerical fit is acceptable, replacing numerical probabilities with epistemic markers that convey the same probabilities can change both the model's choices and the fitted PT parameters.
A fitted PT parameter therefore cannot automatically be treated as a stable measure of the model's risk preference or as evidence that the model makes decisions in the same way as humans.
Our study suggests that researchers should first establish that PT fits the model and that its parameters remain stable across numerical and linguistic representations of uncertainty.
More broadly, when cognitive frameworks developed for humans are applied to foundation models \citep{binz2025foundation}, the wording of a decision problem should therefore be treated as part of the experimental setting because it can systematically shape the model's choices.

\paragraph{On the Validity and Scope of Our Work.}
Our study examines whether PT can reliably describe the choices made by the models we study and whether their choices and fitted PT parameters remain stable when equivalent uncertainty is expressed numerically or linguistically.
Although our experiment includes only 35 lottery pairs, these pairs are not arbitrary. They come from the established experimental economics design of \citet{Tanaka2010RiskMeasure}, which was developed to estimate three components of PT. We sample each pair 256 times, yielding 8,960 choices per condition, and use bootstrap resampling to quantify uncertainty in the estimated PT parameters with narrow 95\% confidence intervals.
Our conclusions apply only to this tightly controlled setting. We do not claim that PT is superior to other decision models, that it applies to all risky choice tasks such as medical diagnosing, that PT itself explains linguistic ambiguity, or that the observed effects are unique to LLMs or demonstrate general differences between LLMs and humans.
A matched human baseline is nontrivial because interpretations of verbal probabilities vary across individuals and contexts \citep{Budescu1995Linguistic,Wallsten1986Probability}, while reproducing our 256-sample-per-lottery protocol at the same scale with human participants would be impractical.
Addressing these broader questions will require comparisons with alternative models, a wider range of decision tasks, and matched human experiments.

\paragraph{On the Implications of Our Work.}
Our findings have several important practical implications for studying and deploying LLMs in risk-sensitive settings.
Researchers should directly verify model fit and parameter stability before interpreting PT parameters as meaningful measures of LLM risk preferences.
In applications where decisions depend on probabilistic information, such as medical or financial decision support, uncertainty expressions should also be carefully and explicitly standardized because semantically similar numerical and linguistic probabilities can induce different choices.
This is particularly important when small linguistic variations can propagate into materially different downstream recommendations.
Finally, although larger models generally exhibit more stable and more PT-like behavior in our experiments, model scale alone does not guarantee fully invariant decision-making under epistemic uncertainty.
This sensitivity also motivates evaluating uncertainty representations explicitly during model validation before deployment.
More on implications are in Appendix~\ref{app:implications}.
\section{Conclusion}
This paper critically re-evaluates the applicability of Prospect Theory (PT) as a descriptive framework for LLM decision-making, particularly under conditions of epistemic uncertainty. Through comprehensive empirical evaluations, our findings reveal three key limitations: (1) Scale Dependency: PT-aligned behavior is not an inherent model capability, but rather emerges only in models with sufficient parameter scale; (2) Representational Divergence: While LLMs demonstrate robust ordinal consistency when ranking epistemic markers, they exhibit severe cross-model divergence in absolute probability mappings; (3) Parameter Instability: The introduction of linguistic uncertainty profoundly destabilizes LLMs, rendering them unable to maintain theoretically coherent, stable, or interpretable PT parameters.
This reveals representational non-invariance.
Ultimately, our study cautions against the uncritical deployment of PT-based frameworks to model or predict LLM behaviors in real-world applications where epistemic ambiguity is ubiquitous. These findings underscore the imperative for future alignment paradigms to move beyond exact numerical probabilities, prioritizing structural robustness and interpretability for autonomous decision-making systems under epistemic uncertainty.
\section*{Limitations}
\label{sec:limitations}

\paragraph{Reasoning configuration effects.}
Our evaluation depends on specific reasoning settings.
For open-source models, we disable reasoning to avoid deterministic expected-value computation, and for closed-source models we impose constraints against explicit expected-value calculation.
This design is aligned with classic economic settings, where participants are asked to give quick and intuitive answers. However, these constraints may artificially limit the analytical capabilities of LLMs.

\paragraph{Generalization to different contexts.} Our work is primarily based on classic economic decision-making context. We focus on this specific setting because it has been widely explored in behavioral economics and offers high interpretability. This approach is also highly relied upon by prior works such as~\cite{Jia2024DMFrame}. While re-framing the decision questions into other practical contexts is an interesting task, it may introduce unexpected confounding influences.

\paragraph{Lack of interpretable parameter patterns.}
We do not observe a theoretically interpretable pattern in prospect-theory parameter changes under epistemic markers.
This lack of regularity likely stems from the inherent ambiguity of epistemic markers, making it difficult to map linguistic uncertainty systematically onto the rigid mathematical structures of PT.
While our findings successfully demonstrate the fragility of PT, which aligns with our primary goal of testing robustness, formulating a unified interpretation for these parameter shifts remains an open question for future research.
\section*{Ethics Statement}
This paper utilize a lottery-based economic questionnaire developed by~\citep{Tanaka2010RiskMeasure} published on  America Economic Association, which allows usage with appropriate citations. The questionaire is done by LLMs, so there is no privacy issues. Our experiment discuss the risk attitude measured under Prospect Theory, which does not contain offensive expressions. The questionnaire is intended to test risk attitude measured by Prospect Theory, and it is used as intended in our paper. 

Our experiment involves the usage of Qwen2.5 series models (7B, 14B, 32B)~\citep{qwen2025qwen25technicalreport} with Apache 2.0 license, Mistral-7B-Instruct-v0.3 with Apache 2.0 license, and Llama3.1-8B-Instruct with Llama3.1 license. They run on an 8x RTX 3090 GPU cluster.

Our paper mainly tests the robustness of PT under epistemic uncertainty, which points out risk of using PT in LLM-related fields. This does not introduce extra risks. Our research focuses on financial decision-making within the English language domain.

\section*{Acknowledgments}
The authors of this paper were supported by the National Key Research and Development Program of China (2025YFE0200500), the ITSP Platform Research Project (ITS/189/23FP) from ITC of Hong Kong, SAR, China, and the AoE (AoE/E-601/24-N), the CRF (No. C6004-25G),  the RIF (R6021-20) and the GRF (16205322) from RGC of Hong Kong, SAR, China. 

\bibliography{custom}

\appendix

\begin{center}
    {\Large\textbf{Appendices}}
\end{center}

\section{Hyperparameters}
\subsection{Probability Mapping Experiment}
\label{app:hyperparameters}
In the probability mapping experiments in \ref{sec:Probability-Mapping-of-Epistemic-Markers}, the monetary reward is fixed at $M=100$. 
This value is chosen to provide a clear and intuitive payoff magnitude without introducing excessive numerical complexity.

The probability parameter $p$ takes values from the set $probs=\{5,15,25,35,45,55,65,75,85,95\}$.
These values are selected to uniformly cover the range of possible probabilities from low to high in increments of 10 percentage points, enabling systematic analysis of internal probability values of epistemic markers.

\subsection{Model Generation}
Temperature is set to $0.7$ for all experiments, following prior work on decision-making behavior of LLMs under uncertain contexts~\citep{Jia2024DMFrame,liu2025costbench,liu2026naacl}. Since our evaluation relies on sampling-based decoding to capture distributional decision behavior, a non-zero temperature is needed. The chosen value balances diversity and coherence and is commonly adopted in LLM evaluations. 
All other decoding and generation hyperparameters use the default settings provided by the HuggingFace implementation.

\begin{table}[h]
\centering
\begin{tabular}{|l|c|}
\hline
\textbf{Hyperparameter} & \textbf{Value} \\
\hline
Temperature & 0.7 \\
\hline
Maximum new tokens & 8 \\
\hline
History length & 10 \\
\hline
Number of lottery rounds & 35 \\
\hline
\end{tabular}
\caption{Key hyperparameters for model generation}
\label{tab:hyperparams}
\end{table}

\subsection{Experiment}
\label{app:experiment-hyperparameters}

Table~\ref{tab:decision-eval-setup-hyperparameters} summarizes the experiment setup hyperparameters used in our decision-making behavior evaluation. 
We sample each lottery question $\eta=256$ times to estimate the empirical choice probability, and evaluate the agent over $M=35$ lottery-choice questions from the three-series design. 
For uncertainty estimation, we report confidence intervals using significance level $\alpha=0.05$, corresponding to a confidence level of $1-\alpha=0.95$.

\begin{table}[t]
\centering
\small
\setlength{\tabcolsep}{3pt}
\begin{tabular}{l p{0.7\linewidth} c}
\toprule
\textbf{Setup} & \textbf{Description} & \textbf{Value} \\
\midrule
$\eta$ & Number of repeated samples per lottery question & $256$ \\
$M$ & Number of lottery-choice questions & $35$ \\
$\alpha$ & Significance level for confidence intervals & $0.05$ \\
$1-\alpha$ & Confidence level & $0.95$ \\
\bottomrule
\end{tabular}
\caption{Experiment setup hyperparameters used in decision-making behavior evaluation.}
\label{tab:decision-eval-setup-hyperparameters}
\end{table}

\section{Lottery Design}
\label{app:prospect-settings}

Table \ref{tab:series1}, \ref{tab:series2} and \ref{tab:series3} shows the lottery design for PT parameter estimation. The values here are from~\cite{Tanaka2010RiskMeasure}. They are specially designed to get best PT parameters.

\section{Prompt Design}
\label{app:prompt-design}
Figure \ref{fig:prompt-design} shows prompt design for decision-making evaluation test and probability mapping test. All lotteries are sampled 256 times. To simulate human decision-making, while keeping the model directly output its answer, up to 15 history decisions are maintained. Meanwhile, we use a random order for the 35 lotteries to relieve positional bias. An introduction is provided at the very beginning. NUMBER will be replaced by the values stated in table \ref{tab:series1}, \ref{tab:series2} and \ref{tab:series3} .

\section{Marker Replacement Rules}
\label{marker-replacement}
For the details of how we replace probabilities with markers, see Table \ref{marker-subrule}.

\begin{table*}[htbp]
\centering
\resizebox{\linewidth}{!}{
\begin{tabular}{|l|c|c|c|c|}

\hline

\textbf{Model} & \textbf{30\%} & \textbf{70\%} & \textbf{10\%} & \textbf{90\%} \\ \hline

Qwen2.5-7B-Instruct & uncertain & almost certain & somewhat likely & highly likely \\ \hline

Llama3.1-8B-Instruct & likely & almost certain & very unlikely & almost certain \\ \hline

Mistral-7B-Instruct-v0.3 & very unlikely & highly likely & highly unlikely & almost certain \\ \hline

Qwen2.5-14B-Instruct & somewhat unlikely & highly likely & very unlikely & almost certain \\ \hline

Qwen2.5-32B-Instruct & somewhat unlikely & probable & somewhat likely & almost certain \\ \hline

\end{tabular}
}
\caption{Marker Replacement Rules for Different Models. This is determined for introducing the least numeric differences, and balancing model and human interpretations.}
\label{marker-subrule}
\end{table*}

\begin{table}[t]
\centering
\begin{tabular}{|c|c c|c c|}
\hline
& \multicolumn{2}{c|}{\textbf{Option K}} & \multicolumn{2}{c|}{\textbf{Option U}} \\
\hline
\textbf{Lottery} & 30\% & 70\% & 10\% & 90\% \\
\hline
1  & 40 & 10 & 68   & 5   \\
2  & 40 & 10 & 75   & 5   \\
3  & 40 & 10 & 83   & 5   \\
4  & 40 & 10 & 93   & 5   \\
5  & 40 & 10 & 106  & 5   \\
6  & 40 & 10 & 125  & 5   \\
7  & 40 & 10 & 150  & 5   \\
8  & 40 & 10 & 185  & 5   \\
9  & 40 & 10 & 220  & 5   \\
10 & 40 & 10 & 300  & 5   \\
11 & 40 & 10 & 400  & 5   \\
12 & 40 & 10 & 600  & 5   \\
13 & 40 & 10 & 1000 & 5   \\
14 & 40 & 10 & 1700 & 5   \\
\hline
\end{tabular}
\caption{Series 1: both options are gains.}
\label{tab:series1}
\end{table}

\begin{table}[t]
\centering
\begin{tabular}{|c|c c|c c|}
\hline
& \multicolumn{2}{c|}{\textbf{Option K}} & \multicolumn{2}{c|}{\textbf{Option U}} \\
\hline
\textbf{Lottery} & 90\% & 10\% & 70\% & 30\% \\
\hline
1  & 40 & 30 & 54   & 5   \\
2  & 40 & 30 & 56   & 5   \\
3  & 40 & 30 & 58   & 5   \\
4  & 40 & 30 & 60   & 5   \\
5  & 40 & 30 & 62   & 5   \\
6  & 40 & 30 & 65   & 5   \\
7  & 40 & 30 & 68   & 5   \\
8  & 40 & 30 & 72   & 5   \\
9  & 40 & 30 & 77   & 5   \\
10 & 40 & 30 & 83   & 5   \\
11 & 40 & 30 & 90   & 5   \\
12 & 40 & 30 & 100  & 5   \\
13 & 40 & 30 & 110  & 5   \\
14 & 40 & 30 & 130  & 5   \\
\hline
\end{tabular}
\caption{Series 2: both options are gains.}
\label{tab:series2}
\end{table}

\begin{table}[t]
\centering
\begin{tabular}{|c|c c|c c|}
\hline
\multicolumn{1}{|c|}{} & \multicolumn{2}{c|}{\textbf{Option K}} & \multicolumn{2}{c|}{\textbf{Option U}} \\
\cline{2-5}
\multicolumn{1}{|c|}{} & \textbf{50\%} & \textbf{50\%} & \textbf{50\%} & \textbf{50\%} \\
\cline{2-5}
\textbf{Lottery} & \textbf{Win} & \textbf{Lose} & \textbf{Win} & \textbf{Lose} \\
\hline
1  & 25 & 4  & 30 & 21 \\
2  & 4  & 4  & 30 & 21 \\
3  & 1  & 4  & 30 & 21 \\
4  & 1  & 4  & 30 & 16 \\
5  & 1  & 8  & 30 & 16 \\
6  & 1  & 8  & 30 & 14 \\
7  & 1  & 8  & 30 & 11 \\
\hline
\end{tabular}
\caption{Series 3 both options have gains and losses.}
\label{tab:series3}
\end{table}

\begin{figure*}[htbp]
\resizebox{1\linewidth}{!}{
\begin{tcolorbox}[
    colback=lossgreen!5!white,
    colframe=frame1,
    title=Prompt Templates,
    fonttitle=\bfseries,
    colbacktitle=lossgreen!50!white,
    coltitle=frame1,
    boxrule=1.5pt,
    arc=5pt,
    boxsep=5pt,
    left=12pt,
    right=12pt,
    top=12pt,
    bottom=12pt
]

\textbf{Beginning Instruction} \\
    You are invited to participate in an experiment.  \\
    Your task is to choose between option K and   \\
    option U for each of the following lotteries.  \\
    Here is lottery \{i\}: \\

\textbf{a) Used in Probability Mapping} \\
    For option K: \\
    p\% probability to win \$100. \\
    For option U: \\
    MARKER to win \$100. \\

\textbf{b) Used in PT Estimation} \\
    For option U: \\
    NUMBER\% probability to win \$NUMBER \\
    NUMBER\% probability to win/lose \$NUMBER \\
    For option K: \\
    NUMBER\% probability to win \$NUMBER \\
    NUMBER\% probability to win/lose \$NUMBER \\

\textbf{End Instruction} \\
    Please DO NOT REASON and \\
    DIRECTLY output your choice, \\
    by \*\*ONLY returning \\
    one of the following two labels: \\
    ``[K]'', ``[U]''. \\
    The answer is: \\
\end{tcolorbox}
}
\caption{\textbf{Templates for prompts used in the probability mapping and Prospect Theory estimation tasks.} The design includes an initial instruction, task-specific lottery descriptions, and a fixed closing instruction to ensure direct model responses without reasoning. }
\label{fig:prompt-design}
\end{figure*}

\section{Detailed Experimental Results}
\label{app:detail-results}

\paragraph{Switching probabilities.}
In the main text, we presented selected key experimental results and visualizations. 
To provide the full probability mappings used in our marker-based experiments, Tables~\ref{tab: top 7 markers} and~\ref{tab: bottom 7 markers} report the switching probabilities for the top and bottom seven epistemic markers across models. 
These results show how each model maps linguistic uncertainty expressions to implicit numerical probabilities, which serve as the basis for replacing numerical probabilities with epistemic markers in the re-measurement experiments.

\paragraph{Prospect Theory parameter estimates and model fit.}
To provide a more comprehensive view of model performance across different rounds, we include in this appendix the full set of parameter estimates and model fit metrics. 
Specifically, Tables~\ref{tab:sigma-merged}, \ref{tab:lambda-merged}, and \ref{tab:gamma-merged} report the estimates of parameters $\sigma$, $\lambda$, and $\gamma$ with their 95\% confidence intervals for each model and round. 
Table~\ref{tab:mae_rsquared} summarizes the models’ mean absolute errors (MAE) and McFadden $R^2$ values across rounds. 
These additional data offer deeper insights into model behavior and the dynamics observed throughout the experiments.

\section{Discussion and Implications}
\label{app:implications}
Our findings reveal challenging difficulties in applying human-centric cognitive frameworks, especially Prospect Theory (PT), to LLM decision-making. Different models display distinct interpretations of epistemic uncertainty markers, leading to divergent decision behaviors. Introducing these markers into the decision-making framework substantially alters LLM choices.

Our results suggest that LLMs may not inherently understand risk in human-like ways; their responses often reflect statistical training artifacts rather than cognitively grounded reasoning. \textbf{We recommend conducting regression analyses and goodness-of-fit tests before applying human cognitive models to LLMs.} 

In real-world applications (e.g., medical diagnosis or financial advice), LLMs may give inconsistent recommendations when probabilistic language varies, posing reliability concerns. \textbf{We recommend establishing consistent standards for expressing uncertainty in LLM-driven decision systems.}

Furthermore, larger LLMs tend to exhibit more PT-like decision behavior, with PT parameters more closely aligned to human estimates. \textbf{We recommend using LLMs with at least 14B parameters when integrating PT into decision-making systems.}

\section{Failure Cases Analysis}
\label{app:failure-case-analysis}
In our initial implementation of the marker mapping experiment, we adopted the set of epistemic markers from Table 6 (“Human Judgements of Templates Based on Reliability”) in~\citep{Zhou2024relying}. 
These markers were originally designed to test both human and LMs’ judgments of the reliability conveyed by these expressions.
However, when directly applied to our economic decision-making setting, the resulting mappings for LLMs were unexpectedly unstable and, in some cases, counterintuitive.

\begin{figure}[t]
     \centering
     \includegraphics[width=1\linewidth]{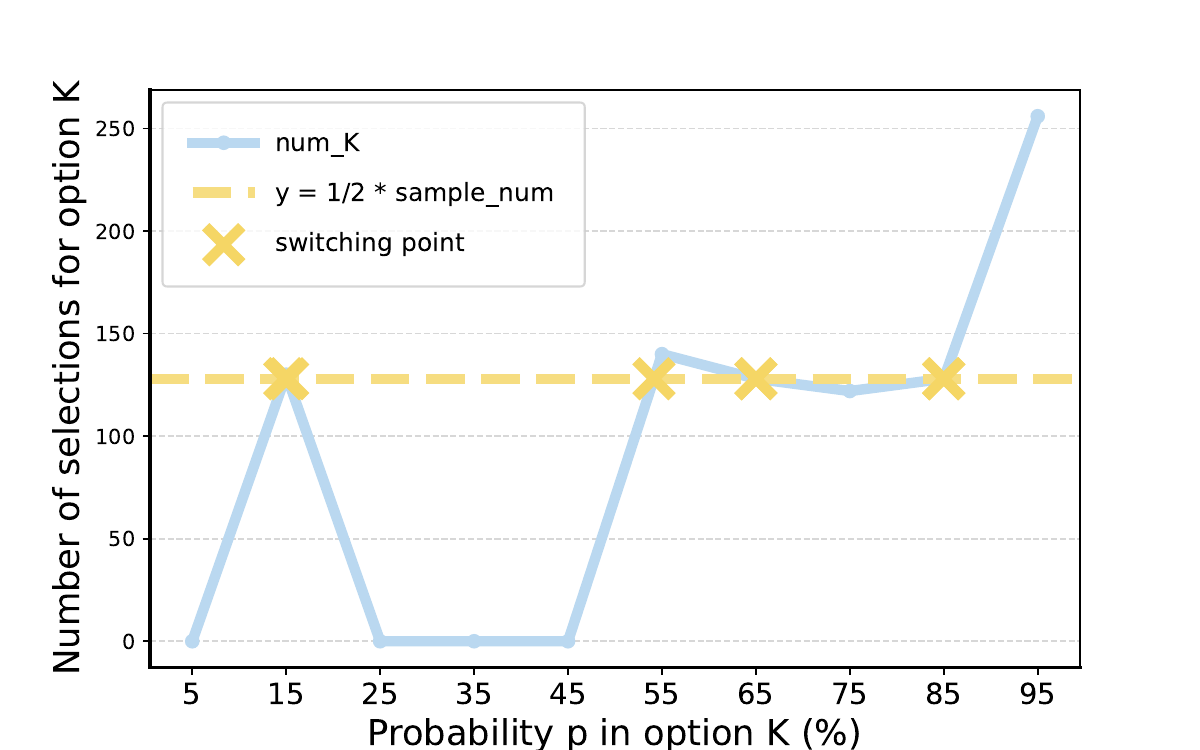}
     \caption{\textbf{An example of non-monotonic result with multiple switching points.} This result comes from marker \textit{``It’s undoubtedly to''} with \textit{Qwen2.5-7B-Instruct} model over 256 trials. The blue line shows counts of option K selections, and the yellow dashed line marks half the trials. The crosses denote switching points.}
    \label{fig:failure}
    \vspace{-0.1in}
\end{figure}

We summarize two major issues observed in the experimental outcomes:

\textbf{(1) Highly oscillatory choice patterns.}
Ideally, the number of times the model selects option K should increase monotonically with $p$, yielding a single and well-defined switching point. 
In practice, the selection curves were often non-monotonic, with multiple apparent switching points, which made the mapping probability $p_{\text{mapping}}$ ill-defined.
An example is shown in Figure \ref{fig:failure}.

\begin{figure*}[h!]
\begin{tcolorbox}[
    colback=riskblue!5!white,
    colframe=frame2,
    title=Failure Case of Markers Mapping,
    fonttitle=\bfseries,
    colbacktitle=riskblue!45!white,
    coltitle=frame2!80!white,
    boxrule=1.5pt,
    arc=5pt,
    boxsep=5pt,
    left=12pt,
    right=12pt,
    top=12pt,
    bottom=12pt
]
\noindent\textbf{user:}
\vspace{-0.2em}
\begin{list}{}{
  \setlength{\leftmargin}{3em}   
  \setlength{\rightmargin}{0pt}
  \setlength{\itemsep}{0pt}
  \setlength{\parsep}{0pt}
  \setlength{\parskip}{0pt}
}
\item
Here is lottery 2:
\vspace{0.4em}
\item \quad \quad For option K: 15\% probability to win \$100.

\item \quad \quad For option U: It's extremely certain to win \$100.

\vspace{0.8em}
Please DO NOT REASON and DIRECTLY output your choice,\\
by ONLY returning one of the following two labels: ``[K]'', ``[U]''.\\[0.8em]
The answer is:
\end{list}

\vspace{0.4em}
\noindent\textcolor{frame2}{\rule{\linewidth}{0.8pt}}
\vspace{0.4em}

\noindent\textbf{assistant:}
\vspace{-0.2em}
\begin{list}{}{
  \setlength{\leftmargin}{3em}
  \setlength{\rightmargin}{0pt}
  \setlength{\itemsep}{0pt}
  \setlength{\parsep}{0pt}
  \setlength{\parskip}{0pt}
}
\item
K
\end{list}

\end{tcolorbox}
\caption{\textbf{An example of mapping high-certainty markers to low probablity.} In this case, \textit{Qwen2.5-7B-Instruct} preferred option K when $p=15\%$ and option U was described using \textit{``It’s extremely certain to''}, implying a far lower internal probability than expected.}
\label{fig:failure-case}
\vspace{-0.1in}
\end{figure*}

\textbf{(2) Severe semantic mismatches for high-certainty markers.}
Some markers, such as \textit{``It's extremely certain to''}, convey very high certainty in human interpretation, but were mapped by the model to surprisingly low numerical probabilities. 
An example is shown in Figure \ref{fig:failure-case}.

We hypothesize that these issues may stem from several factors:

\textbf{(1) Marker length and syntactic complexity.}
The selected markers were often not single words or short phrases, but full clausal structures. 
This may introduce additional semantic and syntactic cues unrelated to uncertainty, thereby interfering with probability interpretation.

\textbf{(2) Shift from first-person to third-person framing.}
The original markers in~\citep{Zhou2024relying} were presented in the first person (e.g., \textit{``I am not confident, maybe it's...''}), whereas our experiment reformulated them into third-person expressions (e.g., \textit{``It's not conﬁdent, maybe can...''}). 

\textbf{(3) Intrinsic instability of epistemic markers.}
Even for human interpretation, such markers are context-dependent and inherently imprecise~\citep{Liu2025markersinconf, multirole-r1,zong2025critical,guo2025mathematical}. 
Their probability mapping by LLMs in economic decision-making contexts may exhibit fundamental reliability flaws~\citep{DBLP:journals/corr/abs-2505-13259,GProofT}.

These limitations motivated the redesign of our marker set and prompt formulation in subsequent experiments.

\section{Comparative Analysis with Prior Work}
\label{app:comparative_analysis}

Prior work by~\cite{horton2023large} suggests that LLMs can simulate economic agents and successfully replicate Prospect Theory (PT), seemingly contradicting our negative results. However, this disparity highlights the critical distinction between \textit{persona-conditioned} and \textit{native} behaviors.~\cite{horton2023large} demonstrates PT emergence primarily under precise numerical probabilities and explicit persona prompting (e.g., instructing the model to act as ``bad at math''). They note that unprompted, capable models default to rational expected-value (EV) computation. To assess native risk attitudes, we explicitly instructed models \textit{not} to calculate EV, revealing that without persona constraints, LLMs inherently lack stable PT adherence.

Our observation of attenuated loss aversion is strongly corroborated by~\cite{yee2026calibrating}, who found that default LLMs exhibit significantly lower loss aversion than human benchmarks unless heavily conditioned. Furthermore, our core contribution identifies \textit{epistemic uncertainty} as a critical breaking point. While LLMs may pass basic numeric risk tests, replacing numbers with ubiquitous linguistic ambiguity (e.g., ``highly likely'') collapses their structural consistency. This aligns with~\cite{liu2025evaluating} and~\cite{scirep2024predict}, who show LLM risk preferences degrade or even reverse in nuanced, real-world contexts~\cite{payne2025analysis}.

\textbf{Implications for System Design.} These negative results serve as a crucial warning for AI deployment. System designers cannot rely on LLMs to natively exhibit human-like caution or risk aversion under ambiguity. For safety-critical applications (e.g., healthcare, finance), epistemic uncertainty must be strictly translated into standardized numeric probabilities, or models must be explicitly aligned via persona-prompting to enforce desired risk profiles.

\begin{table*}[t]
\centering
\resizebox{\textwidth}{!}{
\begin{tabular}{lccccccc}
\toprule
\multirow{2}{*}{\textbf{Model}} & \multicolumn{7}{c}{Top 7 Epistemic Markers} \\
\cmidrule(lr){2-8}
& \textbf{almost certain} & \textbf{highly likely} & \textbf{very likely} & \textbf{likely} & \textbf{probable} & \textbf{somewhat likely} & \textbf{possible} \\
\midrule
\textit{Llama-3.1-8B-Instruct} & 87.92 & 56.04 & 58.00 & 41.80 & 44.23 & 36.71 & 36.29 \\
\textit{Mistral-7B-Instruct-v0.3} & 96.80 & 67.89 & 63.10 & 57.50 & 87.22 & 48.98 & 52.16 \\
\textit{Qwen2.5-7B-Instruct} & 82.71 & 67.00 & 67.06 & \phantom{0}4.78 & \phantom{0}3.44 & \phantom{0}8.93 & \phantom{0}4.51 \\
\textit{Qwen2.5-14B-Instruct} & 91.56 & 55.00 & 54.10 & 42.38 & 26.51 & 32.47 & 38.38 \\
\textit{Qwen2.5-32B-Instruct} & 97.50 & 95.08 & 82.82 & 65.00 & 54.74 & 46.08 & 55.00 \\
\midrule
\textit{GPT-5-Mini} & 97.50 & 70.81 & 67.78 & 50.13 & 48.12 & 44.55 & \phantom{0}4.24 \\
\textit{Gemini-2.5-Flash} & 97.16 & 61.32 & 52.34 & 45.73 & 32.78 & \phantom{0}4.94 & \phantom{0}3.64 \\
\bottomrule
\end{tabular}
}
\caption{\textbf{Switching probabilities (\%) for top 7 epistemic markers.}  
For \textit{GPT-5-Mini} and \textit{Gemini-2.5-Flash}, we allow reasoning while prohibiting expected-value calculation. See \ref{tab: bottom 7 markers} for bottem 7 markers mappings.
}
\label{tab: top 7 markers}
\end{table*}

\begin{table*}[t]
\centering
\resizebox{\textwidth}{!}{
\begin{tabular}{lccccccc}
\toprule
\multirow{2}{*}{\textbf{Model}} & \multicolumn{7}{c}{Bottom 7 Epistemic Markers} \\
\cmidrule(lr){2-8}
& \textbf{uncertain} & \textbf{somewhat unlikely} & \textbf{unlikely} & \textbf{not likely} & \textbf{doubtful} & \textbf{very unlikely} & \textbf{highly unlikely} \\
\midrule
\textit{Llama-3.1-8B-Instruct} & 35.49 & 33.71 & 33.49 & 36.91 & 33.65 & 31.30 & 32.83 \\
\textit{Mistral-7B-Instruct-v0.3} & 48.08 & 40.15 & 38.27 & 30.93 & 34.03 & 29.47 & 27.88 \\
\textit{Qwen2.5-7B-Instruct} & 35.58 & 27.24 & 19.90 & 27.70 & 25.69 & 18.37 & 19.32 \\
\textit{Qwen2.5-14B-Instruct} & 29.03 & 26.45 & 19.10 & 20.82 & 13.04 & 10.94 & 10.52 \\
\textit{Qwen2.5-32B-Instruct} & \phantom{0}2.98 & 21.89 & \phantom{0}3.33 & \phantom{0}3.08 & \phantom{0}3.42 & \phantom{0}2.77 & \phantom{0}2.51 \\
\midrule
\textit{GPT-5-Mini} & \phantom{0}4.16 & \phantom{0}4.78 & \phantom{0}2.93 & \phantom{0}3.17 & \phantom{0}3.52 & \phantom{0}2.57 & \phantom{0}2.58 \\
\textit{Gemini-2.5-Flash} & \phantom{0}3.25 & \phantom{0}2.86 & \phantom{0}2.51 & \phantom{0}2.50 & \phantom{0}2.52 & \phantom{0}2.51 & \phantom{0}2.50 \\
\bottomrule
\end{tabular}
}
\caption{\textbf{Switching probabilities (\%) for bottom 7 epistemic markers.} 
Experiment setup is the same as Table~\ref{tab: top 7 markers}.
}
\label{tab: bottom 7 markers}
\end{table*}

\begin{table*}[t]
\centering
\resizebox{\textwidth}{!}{
\begin{tabular}{lcccccc}
\toprule
\multirow{2}{*}{\textbf{Model}} 
& \multicolumn{5}{c}{\textbf{$\sigma$ (95\% CI)}} \\
\cmidrule(lr){2-6}
& \textbf{\cellcolor{baseline}baseline} 
& \textbf{round1} 
& \textbf{round2} 
& \textbf{round3} 
& \textbf{round4} \\
\midrule
\textit{Llama-3.1-8B-Instruct} 
& \cellcolor{baseline}$0.585 \,(0.578, 0.592)$ 
& $0.605 \,(0.599, 0.612)$ 
& $0.573 \,(0.566, 0.580)$ 
& $0.463 \,(0.455, 0.473)$ 
& $0.593 \,(0.585, 0.602)$ \\
\textit{Mistral-7B-Instruct-v0.3} 
& \cellcolor{baseline}$0.534 \,(0.526, 0.543)$ 
& $0.566 \,(0.558, 0.574)$ 
& $0.571 \,(0.564, 0.580)$ 
& $0.384 \,(0.370, 0.399)$ 
& $0.444 \,(0.431, 0.454)$ \\
\textit{Qwen2.5-7B-Instruct} 
& \cellcolor{baseline}$0.429 \,(0.415, 0.445)$ 
& $0.010 \,(0.010, 0.074)$ 
& $0.250 \,(0.232, 0.275)$ 
& $0.010 \,(0.010, 0.071)$ 
& $0.344 \,(0.332, 0.359)$ \\
\textit{Qwen2.5-14B-Instruct} 
& \cellcolor{baseline}$0.503 \,(0.495, 0.511)$ 
& $0.444 \,(0.434, 0.454)$ 
& $0.409 \,(0.398, 0.421)$ 
& $0.563 \,(0.555, 0.573)$ 
& $0.589 \,(0.581, 0.600)$ \\
\textit{Qwen2.5-32B-Instruct} 
& \cellcolor{baseline}$0.598 \,(0.591, 0.605)$ 
& $0.569 \,(0.561, 0.576)$ 
& $0.564 \,(0.556, 0.572)$ 
& $0.647 \,(0.640, 0.655)$ 
& $0.607 \,(0.599, 0.615)$ \\
\bottomrule
\end{tabular}
}
\caption{$\sigma$ estimates with 95\% confidence intervals across different rounds for each model.}
\label{tab:sigma-merged}
\end{table*}

\begin{table*}[t]
\centering
\resizebox{\textwidth}{!}{
\begin{tabular}{lccccc}
\toprule
\multirow{2}{*}{\textbf{Model}} 
& \multicolumn{5}{c}{\textbf{$\lambda$ (95\% CI)}} \\
\cmidrule(lr){2-6}
& \textbf{\cellcolor{baseline}baseline} & \textbf{round1} & \textbf{round2} & \textbf{round3} & \textbf{round4} \\
\midrule
\textit{Llama-3.1-8B-Instruct} 
& \cellcolor{baseline}$0.010 \,(0.010, 0.125)$ 
& $0.010 \,(0.010, 0.117)$ 
& $0.010 \,(0.010, 0.130)$ 
& $0.010 \,(0.010, 0.168)$ 
& $0.010 \,(0.010, 0.112)$ \\
\textit{Mistral-7B-Instruct-v0.3} 
& \cellcolor{baseline}$0.570 \,(0.453, 0.688)$ 
& $0.010 \,(0.010, 0.132)$ 
& $0.010 \,(0.010, 0.135)$ 
& $2.260 \,(2.060, 2.445)$ 
& $0.267 \,(0.105, 0.414)$ \\
\textit{Qwen2.5-7B-Instruct} 
& \cellcolor{baseline}$0.010 \,(0.010, 0.584)$ 
& $4.000 \,(0.010, 4.000)$ 
& $4.000 \,(3.526, 4.000)$ 
& $4.000 \,(0.010, 4.000)$ 
& $4.000 \,(3.736, 4.000)$ \\
\textit{Qwen2.5-14B-Instruct} 
& \cellcolor{baseline}$1.909 \,(1.801, 2.013)$ 
& $1.919 \,(1.784, 2.070)$ 
& $2.851 \,(2.675, 3.023)$ 
& $2.191 \,(2.094, 2.295)$ 
& $2.531 \,(2.409, 2.648)$ \\
\textit{Qwen2.5-32B-Instruct} 
& \cellcolor{baseline}$1.213 \,(1.133, 1.295)$ 
& $1.340 \,(1.250, 1.423)$ 
& $0.953 \,(0.866, 1.036)$ 
& $2.013 \,(1.945, 2.090)$ 
& $1.815 \,(1.736, 1.905)$ \\
\bottomrule
\end{tabular}
}
\caption{$\lambda$ estimates with 95\% confidence intervals across different rounds for each model.}
\label{tab:lambda-merged}
\end{table*}

\begin{table*}[t]
\centering
\resizebox{\textwidth}{!}{
\begin{tabular}{lccccc}
\toprule
\multirow{2}{*}{\textbf{Model}} 
& \multicolumn{5}{c}{\textbf{$\gamma$ (95\% CI)}} \\
\cmidrule(lr){2-6}
& \textbf{\cellcolor{baseline}baseline} & \textbf{round1} & \textbf{round2} & \textbf{round3} & \textbf{round4} \\
\midrule
\textit{Llama-3.1-8B-Instruct} 
& \cellcolor{baseline}$0.753 \,(0.740, 0.767)$ 
& $0.750 \,(0.737, 0.762)$ 
& $0.755 \,(0.741, 0.768)$ 
& $1.020 \,(0.994, 1.053)$ 
& $1.478 \,(1.443, 1.517)$ \\
\textit{Mistral-7B-Instruct-v0.3} 
& \cellcolor{baseline}$0.577 \,(0.565, 0.590)$ 
& $0.564 \,(0.554, 0.574)$ 
& $0.535 \,(0.527, 0.545)$ 
& $1.413 \,(1.300, 1.564)$ 
& $1.332 \,(1.266, 1.414)$ \\
\textit{Qwen2.5-7B-Instruct} 
& \cellcolor{baseline}$3.645 \,(3.501, 3.783)$ 
& $1.156 \,(0.010, 4.000)$ 
& $0.985 \,(0.870, 1.174)$ 
& $1.035 \,(0.010, 3.498)$ 
& $0.987 \,(0.926, 1.053)$ \\
\textit{Qwen2.5-14B-Instruct} 
& \cellcolor{baseline}$0.896 \,(0.875, 0.919)$ 
& $0.933 \,(0.905, 0.967)$ 
& $0.983 \,(0.944, 1.023)$ 
& $1.455 \,(1.414, 1.503)$ 
& $1.742 \,(1.693, 1.803)$ \\
\textit{Qwen2.5-32B-Instruct} 
& \cellcolor{baseline}$0.867 \,(0.851, 0.884)$ 
& $0.778 \,(0.762, 0.793)$ 
& $0.812 \,(0.797, 0.828)$ 
& $0.614 \,(0.605, 0.623)$ 
& $0.664 \,(0.653, 0.674)$ \\
\bottomrule
\end{tabular}
}
\caption{$\gamma$ estimates with 95\% confidence intervals across different rounds for each model.}
\label{tab:gamma-merged}
\end{table*}

\begin{table}[t]
\centering
\resizebox{\textwidth}{!}{
\begin{tabular}{
    l!{\vrule width 0.4pt\hspace{0.1em}\vrule width 0.4pt}
    ccccc!{\vrule width 0.4pt}
    ccccc
}
\toprule
\multirow{2}{*}{\textbf{Model}} 
& \multicolumn{5}{c!{\vrule width 0.4pt}}{\textbf{MAE}}
& \multicolumn{5}{c}{\textbf{McFadden $R^2$}} \\
\cmidrule(lr){2-6} \cmidrule(lr){7-11}
& \textbf{\cellcolor{baseline}baseline} & \textbf{round1} & \textbf{round2} & \textbf{round3} & \textbf{round4}
& \textbf{\cellcolor{baseline}baseline} & \textbf{round1} & \textbf{round2} & \textbf{round3} & \textbf{round4} \\
\midrule
\textit{Llama-3.1-8B-Instruct} 
& \cellcolor{baseline}0.332 & 0.320 & 0.326 & 0.242 & 0.155
& \cellcolor{baseline}0.092 & 0.121 & 0.090 & 0.082 & 0.335 \\
\textit{Mistral-7B-Instruct-v0.3} 
& \cellcolor{baseline}0.155 & 0.202 & 0.196 & 0.067 & 0.075
& \cellcolor{baseline}0.132 & 0.162 & 0.190 & 0.056 & 0.130 \\
\textit{Qwen2.5-7B-Instruct} 
& \cellcolor{baseline}0.047 & 0.125 & 0.159 & 0.387 & 0.157
& \cellcolor{baseline}0.116 & 0.000 & 0.011 & -0.001 & 0.032 \\
\textit{Qwen2.5-14B-Instruct} 
& \cellcolor{baseline}0.257 & 0.152 & 0.069 & 0.150 & 0.157
& \cellcolor{baseline}0.067 & 0.070 & 0.075 & 0.227 & 0.336 \\
\textit{Qwen2.5-32B-Instruct} 
& \cellcolor{baseline}0.161 & 0.127 & 0.129 & 0.224 & 0.256
& \cellcolor{baseline}0.225 & 0.195 & 0.211 & 0.139 & 0.088 \\
\bottomrule
\end{tabular}
}
\caption{Mean absolute error (MAE) and McFadden $R^2$ across different rounds for each model.}
\label{tab:mae_rsquared}
\end{table}

\end{document}